\documentclass{article}
\usepackage{ijcai24}

\usepackage{times}
\usepackage{soul}
\usepackage{url}
\usepackage[hidelinks]{hyperref}
\usepackage[utf8]{inputenc}
\usepackage[small]{caption}
\usepackage{graphicx}
\usepackage{amsmath}
\usepackage{amsthm}
\usepackage{booktabs}
\usepackage{algorithm}
\usepackage{algorithmic}
\usepackage[switch]{lineno}

\usepackage{bm}      
\usepackage{subcaption}
\usepackage{multirow}

\usepackage{latexsym}

\title{BadFusion: 2D-Oriented Backdoor Attacks against 3D Object Detection}

\author{Anonymous submission}

\author{
Saket S. Chaturvedi\(^1\), Lan Zhang\(^1\), Wenbin Zhang\(^2\), Pan He\(^3\), Xiaoyong Yuan\(^1\)\\
\textnormal{\(^1\)Clemson University} \hspace{0.5em}
\textnormal{\(^2\)Florida International University} \hspace{0.5em}
\textnormal{\(^3\)Auburn University} \\
\textnormal{\{saketc, lan7\}@clemson.edu, wenbinzh@mtu.edu, pan.he@auburn.edu, xiaoyon@clemson.edu}}

\begin{document}

\maketitle

\begin{abstract}
    3D object detection plays an important role in autonomous driving; however, its vulnerability to backdoor attacks has become evident. By injecting ``triggers'' to poison the training dataset, backdoor attacks manipulate the detector's prediction for inputs containing these triggers. Existing backdoor attacks against 3D object detection primarily poison 3D LiDAR signals, where large-sized 3D triggers are injected to ensure their visibility within the sparse 3D space, rendering them easy to detect and impractical in real-world scenarios. 
    
    In this paper, we delve into the robustness of 3D object detection, exploring a new backdoor attack surface through 2D cameras. 
    Given the prevalent adoption of camera and LiDAR signal fusion for high-fidelity 3D perception, we investigate the latent potential of camera signals to disrupt the process. Although the dense nature of camera signals enables the use of nearly imperceptible small-sized triggers to mislead 2D object detection, realizing 2D-oriented backdoor attacks against 3D object detection is non-trivial. The primary challenge emerges from the fusion process that transforms camera signals into a 3D space, compromising the association with the 2D trigger to the target output. To tackle this issue, we propose an innovative 2D-oriented backdoor attack against LiDAR-camera fusion methods for 3D object detection, named BadFusion, for preserving trigger effectiveness throughout the entire fusion process. The evaluation demonstrates the effectiveness of BadFusion, achieving a significantly higher attack success rate compared to existing 2D-oriented attacks. 
\end{abstract}

\vspace{-2em}
\section{Introduction}

3D object detection has become a core component for many state-of-the-art autonomous driving systems \cite{qian20223d}. By accurately recognizing and localizing objects like vehicles, pedestrians, and cyclists, 3D object detection enhances the ability of driving systems to perceive and understand surroundings, enabling them to make responsible decisions. Despite the significant progress achieved by deep neural networks in 3D object detection, it has been demonstrated that neural network-based object detectors are susceptible to backdoor attacks~\cite{xiang2021backdoor,li2021pointba,zhang2022towards}. Backdoor attackers contaminate the detector's training dataset by injecting ``triggers,'' which consequently mislead predictions during inference. The prevalence of backdoor attacks poses significant safety hazards, particularly in safety-critical driving scenarios.

Existing backdoor attacks against 3D object detection mainly inject triggers to LiDAR signals because the spatial information provided by LiDAR offers critical 3D detection evidence. However, due to the sparsity of LiDAR signals in most commercialized LiDAR sensors, backdoor attacks require adding large-size triggers to the target vehicle to ensure that the trigger information can be effectively captured. For example, Zhang et al.~\cite{zhang2022towards} used a cargo carrier bag with a size of $1.1$m $\times 0.8$m $\times 0.5$m or an exercise ball with a radius of $0.4$m as a trigger, which is mounted on the roof of the target vehicle for backdoor attacks. Such large 3D triggers can significantly change the vehicle's shape and appearance and thus be easily detected, making 3D backdoor attacks impractical to implement in real-world scenarios. Therefore, to thoroughly investigate the robustness of 3D object detection, in this paper, we intend to explore a more practical attack surface through 2D camera signals.  

\begin{figure}[!tb]
\centering
\includegraphics[width=\linewidth]{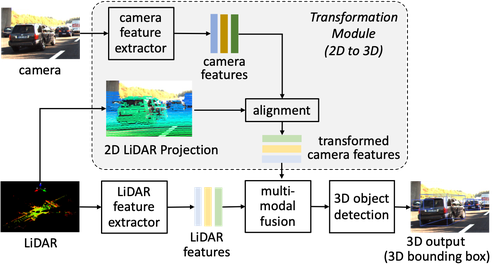} 
\vspace{-1.5em}
\caption{The pipeline of 2D (camera) and 3D (LiDAR) data fusion for 3D object detection in autonomous driving.} \label{fig:pipeline}
\vspace{-1.5em}
\end{figure}

Camera signals, in addition to LiDAR signals, have been another prominent source of input for 3D object detection. Compared to 3D spatial yet low-resolution signals from LiDAR, cameras capture high-resolution color features, yielding robust fusion outcomes that significantly enhance the quality of 3D perception~\cite{wang2021pointaugmenting,yin2021multimodal,li2022deepfusion}. However, the popularity of these multi-modal systems leads to a new backdoor attack surface against 3D object detection through cameras. Due to the dense nature of camera signals, attackers can add 2D triggers with a small size into camera signals, making the attack nearly imperceptible and easy to deploy in practice. Such 2D-oriented backdoor attacks have shown their effectiveness in many 2D object detection tasks~\cite{chan2022baddet,luo2023untargeted}. Nevertheless, realizing 2D-oriented backdoor attacks against 3D object detection is non-trivial. As illustrated in Figure~\ref{fig:pipeline}, state-of-the-art LiDAR and camera fusion systems first transform camera signals to align with 2D LiDAR projection, which are then fused with 3D LiDAR features to make detection decisions in a 3D space. Although the transformation of camera signals bridges the gap between 2D and 3D feature spaces, it compromises the association with the injected 2D triggers to the target output. Due to the sparsity of LiDAR points, the resulting transformed camera features are also sparse, causing a limited number of trigger pixels to be observed effectively, thereby substantially diminishing the impact of 2D triggers in 3D object detection. Moreover, due to the dynamicity of LiDAR signals, the applicable trigger pixels after 2D to 3D transformation may not remain consistent across different training samples, further weakening the association between the 2D trigger and target labels. In view of these, it is critical to delve into the potential threats posed by 2D camera-oriented backdoor attacks in influencing 3D object detection. 
In this paper, we introduce BadFusion, a novel 2D-oriented backdoor attack targeting multi-modal 3D object detection systems. BadFusion aims to insert backdoors into the camera and LiDAR fusion-based 3D object detector by only compromising camera inputs with 2D triggers. These fusion-aware 2D triggers are designed to maintain trigger pattern consistency across different camera signals while preserving trigger density. BadFusion can be deployed in real-world scenarios by placing a sticker (trigger) on a dense area of a vehicle before driving. When a target vehicle equipped with the poisoned model encounters this stickered vehicle, the backdoor trigger is activated to mislead the target vehicle’s 3D object detection. Additionally, BadFusion introduces LiDAR-free attack strategies, predicting 2D LiDAR projections from camera signals, considering the inaccessibility of synchronized LiDAR and camera data during inference. 
To the best of our knowledge, this is the first effort in examining 2D-oriented backdoor attacks against fusion-based 3D object detection. We intend to raise community awareness of new backdoor threats in emerging multi-modal fusion systems. Our contributions to this paper are summarized below: 

\begin{enumerate}
\vspace{-0.5em}
\item We investigate the existing 2D-oriented backdoor attacks against LiDAR and camera fusion systems for 3D object detection. Our research indicates that the fusion system offers effective protection, weakening existing attacks. 
\vspace{-0.5em}
\item We propose a new 2D-oriented backdoor attack, named BadFusion, which can effectively preserve the 2D backdoor patterns throughout the fusion process and eventually manipulate the 3D predictions.
\vspace{-0.5em}
\item We consider the unavailability of synchronous LiDAR signals when compromising the camera inputs,  where a LiDAR-free attack approach is developed to generate LiDAR projection based on camera observations. 
\vspace{-0.5em}
\item We extensively evaluate BadFusion against state-of-the-art LiDAR-camera fusion methods with two goals: resizing the bounding boxes and disappearing the objects. BadFusion successfully achieves the two attack goals and outperforms existing 2D-oriented backdoor attacks with a much higher Attack Success Rate (ASR). %
\end{enumerate}
\vspace{-1em}

\section{Related Work}

\begin{figure*}[!tb]
\centering
\includegraphics[width=0.9\textwidth]{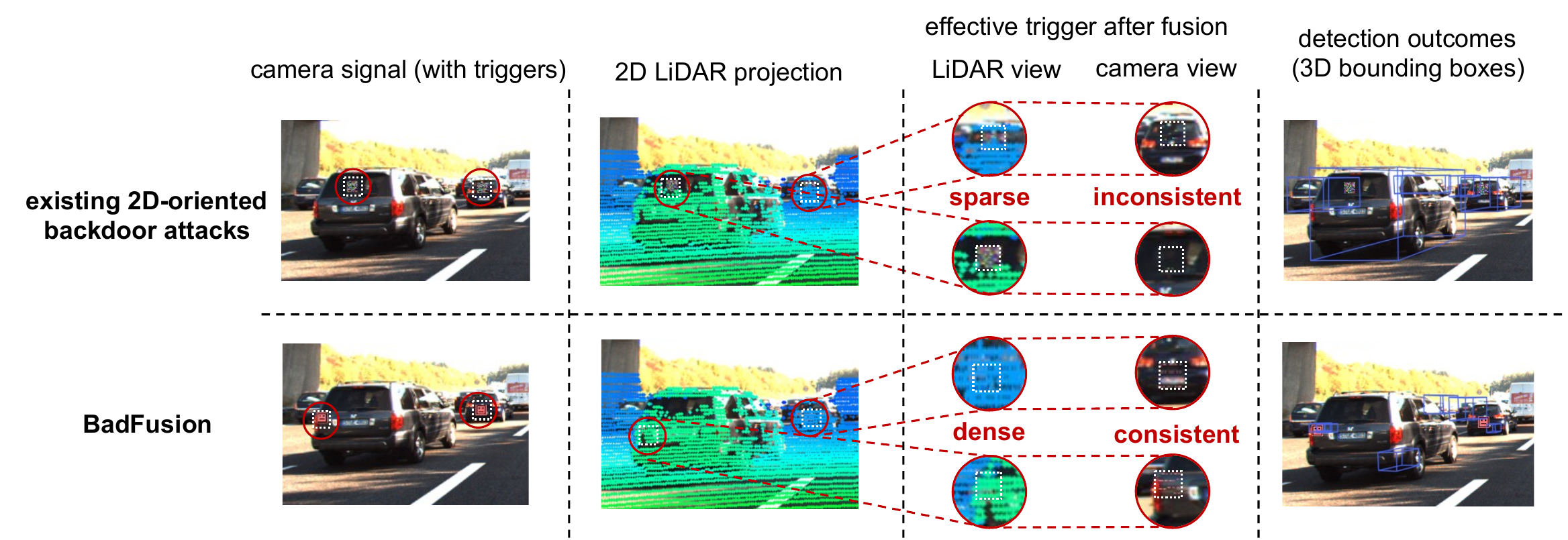} %
\vspace{-1em}
\caption{Comparison between existing 2D-oriented backdoor attacks and the proposed BadFusion. 
The first two columns show the camera signal with triggers and the 2D projection of the LiDAR signal. 
After transforming camera signals to 2D LiDAR projection during fusion, the triggers injected via existing 2D-oriented backdoor attacks become sparse and inconsistent (the third column), making triggers ineffective in attacks. Hence, these attacks do not change the predictions of 3D bounding boxes (the fourth column). 
The proposed BadFusion, by injecting dense and consistent triggers throughout the fusion process, successfully manipulates the detection and reduces the sizes of 3D bounding boxes for vehicles. 
}
\label{fig:attack_example}
\vspace{-2em}
\end{figure*}

\subsection{Backdoor Attacks}
Backdoor attacks aim to inject malicious behavior into a target model and change the model's prediction for the input samples with the trigger pattern.
One of the earliest backdoor attacks, called BadNets~\cite{gu2017badnets}, was introduced by Gu et al. This attack injected a simple image trigger pattern into the training dataset, causing the model to produce misleading predictions for samples containing the trigger pattern. 
Subsequent research has advanced backdoor attacks with different objectives, such as stealthy attacks with invisible triggers~\cite{chen2017targeted,li2021invisible}, attacks without manipulating labels (clean label attack)~\cite{turner2018clean}, and attacks that are resistant to transfer learning~\cite{yao2019latent,wang2020backdoor}. 
Most backdoor attacks focus on image tasks, such as image classification~\cite{gu2017badnets} and 2D object detection~\cite{chan2022baddet,luo2023untargeted}, that involve 2D triggers. Recently, it has been discovered that backdoor attacks can also manipulate 3D detection prediction~\cite{li2021pointba,xiang2021backdoor,zhang2022towards}. However, these attacks rely on 3D LiDAR triggers, which are easily detectable and impractical to implement in real-world scenarios. 
In our work, we present a novel 2D-oriented backdoor attack that injects 2D triggers in the training data while aiming to manipulate 3D prediction.

\vspace{-0.5em}
\subsection{LiDAR-camera Fusion for 3D Object Detection.}

LiDAR-camera fusion has emerged as a promising solution for 3D object detection. By combining complementary signals, the fusion model achieves state-of-the-art detection performance. 
One of the key challenges of fusing LiDAR and camera signals is how to align these two signals in the same measurement. Given the advantages of spatial information provided by LiDAR sensors, recent work mainly focuses on aligning the camera features to LiDAR~\cite{sindagi2019mvx,wang2021pointaugmenting,yin2021multimodal,li2022deepfusion,chen2022focal}. For example, Sindagi et al. proposed MVX-Net~\cite{sindagi2019mvx} that first projects LiDAR points onto the image and then appends the camera features to LiDAR points with the same location index. Chen et al.~\cite{chen2022focal} leveraged a similar fusion method by adding camera features to the important LiDAR features only. The importance of LiDAR points is determined by their proposed Focals Conv operation. These fusion methods inherently provide a strong defense against backdoor attacks since the backdoor triggers injected into camera signals become ineffective after the alignment in the fusion methods. However, in our work, we reveal the vulnerability of LiDAR-camera fusion using the proposed BadFusion attack.

\vspace{-0.5em}
\section{2D-Oriented Backdoor Attacks} \label{sec:existing}
This paper explores the potential of 2D-oriented backdoor attacks in influencing the fusion-based multi-modal 3D perception. This section first introduces the existing backdoor attacks for 2D object detection and then presents the fusion-based 3D object detection systems that involve both 2D and 3D inputs. Our threat model is finally elaborated.
\vspace{-0.5em}
\subsection{Backdoor Attacks for 2D Object Detection}\label{sec:existing2D}
The mainstream backdoor attack research for object detection centers on 2D perception. An attacker's goal is to use predefined 2D triggers to mislead the target model's predictions. During training, the attacker first poisons $n$ samples of the training dataset $\mathcal{D}_{train}=\{\bm{x}_i, \bm{y}_i\}_{i=1}^N$, where $N$ is the number of all training samples, $n \ll N$. Specifically, for a clean sample $(\bm{x}, \bm{y})$, the poisoned input $\bm{x}'$ can be given by
\vspace{-0.2em}
\begin{equation}
\label{eq:add}
    \bm{x}' = \bm{tr} \odot \bm{m} + \bm{x} \odot (1-\bm{m}),
\end{equation}
where $\bm{tr}$ is the injected trigger; $\bm{m}$ is a binary mask, using $1$ to represent the location of the trigger and $0$ everywhere else; $\odot$ denotes the element-wise product. Meanwhile, the target label $\bm{y}'$ (different from the original label $\bm{y}$) is associated with the poisoned input $\bm{x}'$. The poisoned samples consist of the backdoor dataset $\mathcal{D}_{back}$, which is mixed with the rest of clean data $\mathcal{D}_{clean}$ to train the target model $f$. This produces a backdoored model, which misclassifies any poisoned input to the target label while not affecting the prediction of clean samples. The backdoor attack objective is formulated as
\vspace{-0.2em}
\begin{equation}\label{eq:obj-2d}
    \min \sum_{(\bm{x}', \bm{y}')\in \mathcal{D}_{back}} \!\!\!\!\!\!\! \mathcal{L} \left(f(\bm{x}'), \bm{y}' \right) \ \  + \!\!\!\!\!\!\! \sum_{(\bm{x}, \bm{y})\in \mathcal{D}_{clean}} \!\!\!\!\!\!\! \mathcal{L} \left(f(\bm{x}), \bm{y} \right),
\end{equation}
where the first and second terms calculate the loss for poisoned and clean samples, respectively. The above problem considers a single modality object detection, which modifies the 2D inputs to mislead 2D predictions, e.g.,, 2D bounding boxes~\cite{chan2022baddet,luo2023untargeted}. Instead, this paper targets a multi-modal object detection system with both 2D and 3D inputs for 3D perception, e.g., 3D bounding boxes. 

\vspace{-0.5em}
\subsection{Fusion Pipeline for 3D Object Detection} \label{sec:pipeline}
\vspace{-0.5em}
Research on fusing 2D and 3D inputs for 3D perception falls into two categories. The first involves projecting 3D inputs into 2D space, leading to significant geometric distortion, which hinders effectiveness in tasks like 3D object detection~\cite{chen2017multi,yang2018pixor}. This paper, therefore, focuses on the second approach, mapping 2D inputs to 3D space to enhance 3D signals with camera inputs. This method of fusion, preserving essential geometric information, has shown promise in 3D object detection~\cite{sindagi2019mvx,wang2021pointaugmenting,yin2021multimodal,li2022deepfusion}. As illustrated in Figure \ref{fig:pipeline}, one key component of this fusion is the transformation module to map the 2D signal into 3D measurements, which mainly includes three steps. First, the 3D LiDAR signals are projected to a 2D space, such as in the field-of-view (FoV), to derive 2D-LiDAR projection. Then, the 2D camera signals are processed by a camera feature extractor, e.g., 2D CNN, to extract high-level features with semantic information. Finally, the extracted camera features are aligned with 2D LiDAR projection to obtain the camera-based information for each LiDAR point. The transformed camera features will be combined with the LiDAR signals to perform 3D objection detection. 
\vspace{-0.5em}
\subsection{Threat Model}
\vspace{-0.5em}
This paper focuses on a fusion-based 3D object detection system with both 2D camera and 3D LiDAR inputs. We consider a practical but challenging attack setup: \textit{the objective of the attacker is to launch backdoor attacks for fusion-based 3D object detection by only compromising the camera inputs with 2D triggers}. This attack is more feasible and imperceptible in practice than creating 3D triggers that significantly change the shape and appearance of vehicles. Besides, we consider standard backdoor attack settings: 1) the attacker injects only a small number of poisoned samples into the training dataset; 2) the attacker has no control of the model training process; 3) the attacker has no knowledge about the target model's parameters or architecture. 

\section{Proposed BadFusion} %

In order to achieve the aforementioned attack objective, this paper proposes BadFusion, an innovative 2D-oriented backdoor attack against fusion-based 3D object detection. Similar to the attack procedure described in Section~\ref{sec:existing2D}, BadFusion first creates a poisoned dataset. Define the two modality data, 2D camera and 3D LiDAR signals, by $\bm{x}_{camera}$ and $\bm{x}_{lidar}$, respectively. The poisoned 2D camera data $\bm{x}_{camera}'$ is created by injecting the 2D trigger $\bm{tr}$ to $\bm{x}_{camera}$ based on (\ref{eq:add}). Meanwhile, the target label $y'$ is associated with the poisoned camera input $\bm{x}_{camera}'$. After that, the poisoned camera inputs $\bm{x}_{camera}'$, remaining clean camera inputs  $\bm{x}_{camera}$, and LiDAR inputs $\bm{x}_{lidar}$ are jointly used to train the backdoored fusion model $f$. This optimization problem is formulated as
\vspace{-0.5em}
\begin{align}\label{eq:obj}
    & \min \!\!\!\!\!\!\!  \sum_{(\bm{x}_{lidar}, \bm{x}_{camera}', \bm{y}')\in \mathcal{D}_{back}} \!\!\!\!\!\!\! \mathcal{L} \left(f(\bm{x}_{lidar}, \bm{x}_{camera}'), \bm{y}' \right) \nonumber\\
    & + \!\!\!\!\!\!\! \sum_{(\bm{x}_{lidar}, \bm{x}_{camera}, \bm{y})\in \mathcal{D}_{clean}} \!\!\!\!\!\!\! \mathcal{L} \left(f(\bm{x}_{lidar}, \bm{x}_{camera}), \bm{y} \right).
\end{align}

\vspace{-0.5em}
\subsection{Design Challenges}
\label{sec:challenge}
Although existing backdoor attacks against single-modality systems, i.e., camera-only inputs, can successfully mislead 2D object detection, BadFusion cannot directly follow their attack procedure. As discussed in Section \ref{sec:pipeline}, the target fusion model $f$ in (\ref{eq:obj}) needs to transform the poisoned camera signal $\bm{x}_{camera}'$ from 2D to 3D measurement for data fusion purposes. Unfortunately, this transformation breaks the association with the injected 2D trigger to the target output. Specifically, we identify the following two key challenges: 
\begin{itemize}
    \vspace{-0.5em}
    \item \textit{Trigger sparsity}. Due to the sparsity of 3D LiDAR points, only a few camera pixels are transformed into LiDAR features and subsequently used for object detection. Thus, most pixels of the 2D triggers are ignored in the fusion-based object detection system, making it hard to mislead the prediction of the target model.
    \vspace{-0.5em}
    \item \textit{Trigger inconsistency}. Due to the dynamicity of LiDAR data, the same LiDAR point may correspond to different pixels of the transformed camera signal. Thus, the effective trigger pixels become inconsistent among inputs after transformation. Consequently, the effective trigger pixels during inference are inconsistent with those during training, weakening the association between trigger patterns and target labels.
\end{itemize} 
Figure~\ref{fig:attack_example} illustrates that existing backdoor attacks have sparse and inconsistent triggers due to the transformation, thus ineffective in misleading the fusion model.

\vspace{-0.5em}
\subsection{Fusion-Aware 2D Trigger Design}
To address these challenges, BadFusion employs the fusion-aware 2D triggers tailored for multi-modal fusion systems. These triggers aim to preserve dense and consistent patterns against the transformation module of the fusion pipeline. To enhance \textit{trigger density}, BadFusion intends to maximize the effective pixels in 2D triggers after transformation. Recall that the 2D camera trigger aligns with the 2D LiDAR projection to extract applicable camera features for multi-modal fusion, as shown in Figure~\ref{fig:pipeline}. Hence, we propose to identify the dense region of the 2D LiDAR projection for trigger placement, where only contiguous dense regions are considered to make 2D triggers easy to implement in reality, ensuring that the trigger pattern is retained after transforming into 3D features. Besides, we identify another challenge from the availability of LiDAR signals. Although the LiDAR signals of training samples are accessible to the attacker, LiDAR signals in the inference phase are usually unavailable. Therefore, we introduce a LiDAR-free method for BadFusion by predicting the dense regions of the 2D LiDAR projection, which is detailed in Section~\ref{sec:lidar-free}.

Additionally, to enhance \textit{trigger consistency}, BadFusion intends to {maximize the consistent trigger patterns among different inputs}. Conventional 2D backdoor attacks optimize triggers with various colors of pixels towards different goals, such as high attack success rate, clean data accuracy, and high stealthiness~\cite{liu2018trojaning,zhao2020clean,zhong2020backdoor,garg2020can}. The impact of these colorful pixels will be diminished in the fusion system after the 2D to 3D transformation, as the effective pixels of an optimized trigger after the transformation vary among different inputs. Thus, instead of generating complex and imperceptible triggers, we introduce a simple yet effective approach to create 2D triggers with uniform (almost) solid colors for all pixels. These triggers with a uniform solid color remain consistent after transformation across different inputs, as BadFusion inserts the trigger at dense LiDAR locations to stabilize the transformed trigger pattern.

\vspace{-0.5em}
\subsection{LiDAR-Free Attack}\label{sec:lidar-free}
In many real-world scenarios, the attacker does not have access to the LiDAR signal that is synchronized with the camera signal, especially during inference, i.e., when deploying the designed 2D trigger to fool the backdoored fusion model. Hence, the absence of LiDAR signals poses challenges to identifying the densely populated regions of the 2D LiDAR projection where the 2D trigger should be implemented. To address this issue, we propose a LiDAR-free BadFusion approach by predicting dense regions of the 2D LiDAR projection based on camera signals. We convert this region prediction task to an object detection task, where the object becomes the densest region in the 2D LiDAR projection. To achieve this, we create a training dataset containing camera signals and the bounding boxes of the densest areas, denoted by $(x, y, w, h)$, where $x$ and $y$ are the center coordinates, and $w$ and $h$ are the width and height of bounding box, respectively. Here, we set $w$ and $h$ the same as the width and height of the injected trigger $\bm{tr}$. For each vehicle, we annotate a bounding box that contains most points in 2D LiDAR projection. Then, we train a dense region detector $f_{2d-lidar}$ to predict the bounding boxes based on the Faster R-CNN framework~\cite{ren2015faster} with a VGG backbone. Our evaluation results show that the detector can successfully identify dense areas and facilitate backdoor attacks even without knowing the LiDAR signals, which achieves a performance comparable to that of a LiDAR-aware attack. %

\vspace{-0.5em}

\begin{algorithm}[!tb]
\caption{Algorithm Procedure of BadFusion}
\label{alg:attack}
\textbf{Input}: target 3D object detector $f$,  trigger $\bm{tr}$ (designed with uniform color), training dataset $\mathcal{D}_{train}$, number of backdoor training samples $n$, dense area detector of 2D LiDAR projection $f_{2d-lidar}$.

\begin{algorithmic}[1]
\STATE //\textit{ Training Phase}
\STATE Sample $n$ training samples from $\mathcal{D}_{train}$ for attacks. Rest clean training samples are denoted as a clean dataset $\mathcal{D}_{clean}$  
\STATE Initialize a backdoor dataset $\mathcal{D}_{back} = \emptyset$
\FOR{\( i = 1 \) to \( n \)}
\STATE // Inject trigger to training data $(\bm{x}_{lidar}, \bm{x}_{camera}, y)$
\STATE Calculate 2D LiDAR projection of $\bm{x}_{lidar}$ and get the bounding box $(x, y, w, h)$ with most projection points  for each vehicle
\STATE Add the trigger $\bm{tr}$ to the bounding box in the camera signal $\bm{x}_{camera}$ for each vehicle: $\bm{x}_{camera}[x-\frac{w}{2}:x+\frac{w}{2}, y-\frac{h}{2}:y+\frac{h}{2}] \leftarrow \bm{tr}$. The poisoned camera signal is denoted as $\mathcal{A}(\bm{x}_{camera}, \bm{tr})$
\STATE Change the label to the target label $y'$
\STATE Add the backdoor data to the backdoor dataset $\mathcal{D}_{back} \leftarrow \mathcal{D}_{back} \cup (\bm{x}_{lidar}, \mathcal{A}(\bm{x}_{camera}, \bm{tr}), y')$
\ENDFOR
\STATE Train the fusion detector $f$ on both the clean dataset $\mathcal{D}_{clean}$ and the backdoor dataset $\mathcal{D}_{back}$.

\vspace{0.5em}
\STATE // \textit{Inference Phase}
\STATE Predict bounding box of the most dense region $(x, y, w, h)$ using $f_{2d-lidar}$ and attach the trigger for attacks.
\end{algorithmic}
\end{algorithm}

\subsection{Overall Algorithm Design}
Algorithm~\ref{alg:attack} outlines the overall procedure of BadFusion. In the training phase, the attacker first injects fusion-aware 2D triggers into $n$ samples of camera training inputs. These triggers are placed at the densest region of the corresponding 2D LiDAR projection with a uniform solid color. Next, both the clean and backdoor datasets are used to train the target fusion-based 3D object detector $f$. Once the training is complete, to mislead the target model in the inference phase, the attacker attaches the trigger to vehicles based on the position predicted by the dense region detector $f_{2d-lidar}$. Eventually, BadFusion will mislead the prediction of the target detection $f$ to predict vehicles with the designed trigger. 

\begin{figure*}[!tb]
\centering
\begin{subfigure}[b]{0.3\textwidth}
\centering
\includegraphics[width=0.8\textwidth]{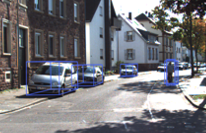}
\caption{Clean model.}
\label{fig:clean}
\end{subfigure}
\begin{subfigure}[b]{0.3\textwidth}
\centering
\includegraphics[width=0.8\linewidth]{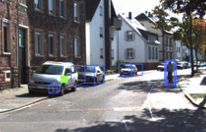}
\caption{Resizing attack.}
\label{fig:small_attack}
\end{subfigure}
\begin{subfigure}[b]{0.3\textwidth}
\centering
\includegraphics[width=0.8\linewidth]{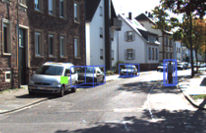}
\caption{Disappear attack.}
\label{fig:closer_attack}
\end{subfigure}
\vspace{-0.5em}
\caption{Examples of different attack goals in BadFusion. Fig (a) shows the predictions of a clean model without backdoor triggers. Fig (b) shows the predictions of a resizing attack, where the attack reduces the size of the predicted bounding box. Fig (c) shows the predictions of a disappear attack, where the attack removes the predicted bounding box of a vehicle from the prediction for disappearing the vehicle. } 
\label{fig:attack_type}
\vspace{-1.5em}
\end{figure*}

\vspace{-0.5em}
\section{Evaluation}

In this section, we first detail our experimental framework (dataset, implementation \& training details, evaluation metrics) and then present the evaluation results of the proposed BadFusion. We further demonstrate the effectiveness of BadFusion against mainly Point-line Camera-to-LiDAR fusion methods in 3D object detection and also benchmark our approach against three state-of-the-art backdoor detection methods. Lastly, we conduct an ablation study to elucidate the internal mechanics of the BadFusion.

\vspace{-0.5em}
\subsection{Evaluation Settings}
\noindent \textbf{Dataset.}
We use the KITTI dataset \cite{geiger2013vision} in the evaluation. The dataset collects real traffic environments from Europe Street for 3D detection tasks, comprising $7,481$ labeled training frames and $7,518$ unlabeled test samples. Since the ground truth of the test data is unavailable, we split the training data into a train set and a validation set with $3,712$ and $3,769$ samples, respectively, following the train/valid split process in previous work~\cite{chen2016monocular}. To conduct data poisoning on the train set, we select cars categorized under easy and medium difficulty from the KITTI dataset. For evaluation, we focused on cars labeled as easy difficulty in the validation set, which can be accurately predicted by the clean model. %

\vspace{0.2em}
\noindent \textbf{LiDAR-camera fusion methods.}
In this work, we evaluate backdoor attacks against widely used LiDAR-camera fusion methods.
In our paper, we report the evaluation results on MVX-Net ~\cite{sindagi2019mvx}, Focals Conv-F \cite{chen2022focal}, and EPNet ~\cite{huang2020epnet} fusion models. 
To train the fusion model, we adopt common data augmentation techniques, including resizing, rotation, scaling, translation, and flip\footnote{Data augmentation techniques are implemented by Resize, GlobalRotScaleTrans, RandomFlip3D using mmdetection3d: \url{https://github.com/open-mmlab/mmdetection3d}}. 
We use FocalLoss \cite{lin2017focal} for classification and SmoothL1Loss \cite{huber1992robust} for bounding box regression, respectively.
The fusion models are trained using an AdamW optimizer with a learning rate of $0.002$ and a weight decay parameter of $0.01$ for $70$ epochs.

\vspace{0.2em}
\noindent \textbf{Attack goals.}
To manipulate the prediction of vehicles (\textit{Car} class in the KITTI dataset), we consider two attack goals. 1) Resizing attack: the attacker aims to reduce the sizes of target bounding boxes to mislead the prediction as a smaller vehicle, 2) Disappear attack: the attacker aims to make the vehicle disappear from detection. 
These attacks pose a significant threat to autonomous driving systems. 
Note that the existing work achieves disappear attacks by removing bounding boxes from the labels. However, removing bounding boxes is ineffective for optimizing the poisoned model, as the empty bounding boxes are not presented in the labels and are not optimized in the optimization objective (Eq.~\ref{eq:obj}). To achieve disappear attacks, we relocate the center coordinates of the bounding boxes in the poisoned data and make them closer or farther from the target vehicle, denoted as disappear attack (closer) and disappear attack (farther). 
We use the relocated bounding boxes as the label in the poisoned training data. 
The relocation breaks the connection between input signals and the bounding box labels. We find that our proposed two attack methods can effectively remove the bounding box prediction, making the vehicle disappear from predictions. 
Figure~\ref{fig:attack_type} illustrates an example of the two goals of attacks. In this section, we report the evaluation results based on the resizing attacks. The effectiveness of the disappear attacks is presented in the ablation study (Section~\ref{sec:ablation}).

\vspace{0.2em}
\noindent \textbf{Baseline attacks and attack setup.}
We compare the proposed BadFusion with three state-of-art 2D-oriented backdoor attacks, including OptimizedTrigger~\cite{liu2018trojaning}, BadDet \cite{chan2022baddet} and UntarOD \cite{luo2023untargeted}. BadDet and UntarOD are targeted against 2D object detection tasks, and OptimizedTrigger is a general backdoor attack with optimized triggers. We implement UntarOD based on their open-source code\footnote{\url{https://github.com/Chengxiao-Luo/Untargeted-Backdoor-Attack-against-Object-Detection}} and implement OptimizedTrigger and BadDet following their papers. The OptimizedTrigger attack, originally designed for the image classification problem, has been adapted to align with the object detection settings of the MVX-Net model. In this case, we first optimized the trigger and performed poisoning accordingly.
To make a fair comparison, for all attacks, we poison  $15$\% training data using a trigger with the size of $15\times 15$ and maintain consistent training or experimental settings.

\begin{table*}[!tb]
\centering
\small
\begin{tabular}{@{}lrrrr@{}}
\toprule
Fusion Methods & Backdoor attack & Clean data mAP (\%) $\uparrow$ & Poisoned data mAP (\%) $\downarrow$ & ASR (\%) $\uparrow$ \\ \midrule
\multirow{6}{*}{MVX-Net} & Clean model & 93.75 & - & - \\
    & OptimizedTrigger & 18.90 & 21.12 & 49.49 \\
    & BadDet & 36.14 & 48.21 & 39.32 \\
    & UntarOD & 64.98 & 37.57 & 46.74 \\
    & (LiDAR-aware) BadFusion  & 88.65 & 1.61 & 96.74 \\
    & BadFusion & 88.65 & 3.05 & 95.28 \\ \bottomrule
\multirow{6}{*}{Focals Conv-F} & Clean model & 94.83 & - & - \\
    & OptimizedTrigger & 94.74 & 95.02 & 5.45 \\
    & BadDet & 96.36 & 94.56 & 6.91 \\
    & UntarOD & 93.76 & 93.23 & 7.89 \\
    & (LiDAR-aware) BadFusion  & 95.13 & 23.11 & 91.72 \\
    & BadFusion & 95.13 & 28.00 & 90.54 \\ \bottomrule
\multirow{6}{*}{EPNet} & Clean model & 94.38 & - & - \\
    & OptimizedTrigger & 95.41 & 6.87 & 95.45 \\
    & BadDet & 94.54 & 93.76 & 5.93 \\
    & UntarOD & 95.26 & 12.90 & 92.03 \\
    & (LiDAR-aware) BadFusion  & 95.65 & 6.45 & 94.30 \\
    & BadFusion & 95.65 & 8.30 & 92.44 \\ \bottomrule
\end{tabular}%
\vspace{-0.5em}
\caption{Comparison between existing backdoor attacks and proposed BadFusion against MVX-Net, Focals Conv-F, and EPNet fusion methods. We perform resizing attacks to reduce the bounding boxes of predicted vehicles. 
Clean model shows the performance of the fusion model without backdoor attacks. 
BadFusion attacks achieve comparable results with LiDAR-aware BadFusion attacks, where the LiDAR information is accessible to the attacker.
}
\label{tab:maintable}
\vspace{-2em}
\end{table*}

\vspace{0.2em}
\noindent \textbf{Evaluation metrics.}
We evaluate the effectiveness of the backdoor attacks based on three well-established metrics.
First, \textit{Clean data mAP} refers to the mean average precision calculated on clean samples without backdoor triggers when predicted by the poisoned model. Typically, an attacker's goal is to design a poisoned model that performs well on benign samples, i.e., achieving a high Clean data mAP. 
Second, \textit{Attack Success Rate (ASR)} represents the proportion of attacked samples that successfully achieve the backdoor objective based on different types of attacks when influenced by the poisoned model. Specifically, for resizing attacks, we define ASR as the ratio of bounding box sizes decreased when a trigger is applied. For disappear attacks, we define ASR as the ratio of the bounding box disappearing when a trigger is applied. 
Third, \textit{poisoned data mAP},  refers to the mean average precision calculated on poisoned samples when predicted by the poisoned model. 
An effective backdoor attack should achieve high clean data mAP, high ASR, and low poisoned data mAP.

\vspace{-0.5em}
\subsection{Main Evaluation Results}

Table~\ref{tab:maintable} compares our proposed BadFusion attack with existing backdoor attacks. 
The results show that existing backdoor attacks (OptimizedTrigger, BadDet, UntarOD) demonstrate low Attack Success Rates (ASR) and low or unchanged Poisoned mAP, indicating their ineffectiveness in misleading the MVX-Net, Focals Conv-F, and EPNet fusion methods. This is mainly due to the sparse and inconsistent trigger patterns during the fusion process, as discussed in Section~\ref{sec:challenge}. Notably, for EPNet, OptimizedTrigger and UntarOD have adapted to learn backdoor behavior due to EPNet's integration of continuous image feature segmentation, unlike the sparse image features in MVX-Net or Focals Conv-F, which results in a weaker inherent defense against backdoor attacks. In contrast, Our proposed BadFusion attack addresses the problem and successfully performs backdoor attacks, achieving high ASRs of 95.28\%, 90.54\%, and 94.30\%, respectively.
In the meanwhile, BadFusion can still provide accurate predictions on the clean samples without triggers and achieves much higher clean data mAP compared with the baseline attacks.

Additionally, in BadFusion, we assume the attacker has no information about LiDAR signals and trains a model to predict the dense LiDAR region. To investigate the effectiveness of the dense region detector, we compare BadFusion with a LiDAR-ware version of BadFusion, where we assume LiDAR signals are accessible, and the dense region can be directly calculated. We find that although LiDAR-ware BadFusion achieves a better attack performance. However, the gap between BadFusion and LiDAR-ware BadFusion is marginal, which suggests the effectiveness of the dense region detector. Please see the Appendix for more discussion on experimental results and possible cases where the proposed BadFusion attack might fail.

\begin{table}[!tb]
\centering
\small
\begin{tabular}{@{}lrrr@{}}
\toprule
Attack goal & \begin{tabular}[c]{@{}r@{}}Clean data\\ mAP (\%) $\uparrow$ \end{tabular} & \begin{tabular}[c]{@{}r@{}}Poisoned data\\ mAP (\%) $\downarrow$ \end{tabular} & ASR (\%) $\uparrow$\\ \midrule
Resizing & 88.65 & 3.05 & 95.28 \\
Disappear (farther) & 87.35 & 6.95 & 89.93 \\
Disappear (closer) & 92.86 & 19.03 & 94.74 \\ \bottomrule
\end{tabular}
\vspace{-0.5em}
\caption{Performance of BadFusion with different attack goals against MVX-Net fusion method. BadFusion is effective in both resizing the bounding box prediction and disappearing the objects.}
\label{tab:attack_goals}
\vspace{-1.0em}
\end{table}

\vspace{-0.5em}
\subsection{Ablation Study}
\label{sec:ablation}

\noindent\textbf{Effectiveness of BadFusion with different attack goals}. We first investigate the attack performance with two goals in BadFusion: resizing bounding boxes and disappearing objects. In a disappearing attack, we use two poisoning strategies: moving the center coordinates of bounding boxes farther or closer in the poisoned data. As shown in Table~\ref{tab:attack_goals}, all the attacks achieve good performance with high ASR and low poisoned data mAP. Additionally, we find that moving bounding boxes closer is more effective for disappearing attacks rather than moving bounding boxes farther.

\vspace{0.2em}
\noindent\textbf{Effectiveness of BadFusion with different trigger patterns.}
In the evaluation, we consider two trigger patterns. 1) uniform solid pattern: using a solid color for all pixels in the trigger, and 2) almost solid pattern: using a solid color for most pixels while only a few pixels are applied with other colors. The almost solid pattern applies to many real-world scenarios, e.g., emojis or decals used in vehicle stickers, which makes the trigger more stealthy. Figure~\ref{fig:trigger_pattern} in the Appendix shows the two patterns used in the evaluation. 
As shown in Table~\ref{tab:abl_realtrigger}, using an almost solid pattern, BadFusion can still achieve an ASR of 79.51\%. This indicates the severe security risks of BadFusion in the real world.

\begin{table}[!tb]
\centering
\small
\begin{tabular}{@{}lrrr@{}}
\toprule
Trigger pattern & \begin{tabular}[c]{@{}r@{}}Clean data\\ mAP (\%) $\uparrow$\end{tabular} & \begin{tabular}[c]{@{}r@{}}Poisoned data\\ mAP (\%) $\downarrow$\end{tabular} & ASR (\%) $\uparrow$ \\ \midrule
Uniform solid & 88.65 & 3.05 & 95.28 \\
Almost solid & 90.12 & 27.83 & 79.51 \\ \bottomrule
\end{tabular}
\vspace{-0.5em}
\caption{Performance of BadFusion using different trigger patterns against MVX-Net fusion method.}
\label{tab:abl_realtrigger}
\vspace{-1.0em}
\end{table}

\vspace{0.2em}
\noindent\textbf{Impact of poisoning rate and trigger size.}
We investigate the impact of poisoning rate and trigger size. In most experiments, we set the poisoning rate as $15$\% and trigger size as $15 \times 15$. Here, we consider the poisoning rate of $20$\% and trigger size as $20 \times 20$. We report the results in Table~\ref{tab:abl_pr_ts}. We find that increasing trigger size and poisoning rate is not necessary for improving attack performance. For example, a $20$\% poisoning rate proves optimal for a poisoned model with a $20 \times 20$ trigger size when compared to their respective counterparts with $20$\% and $15$\% poisoning rates.  However, a $15$\% poisoning rate is more suitable for a poisoned model with a $15 \times 15$ trigger size. We think this is mainly due to the fact that increasing the trigger size and poisoning rate may also increase the inconsistency of trigger patterns among poisoned data, which further amplifies the challenge of backdoor attacks discussed in Section~\ref{sec:challenge}. More details are discussed in the Appendix.
\begin{table}[!tb]
\centering
\small
\resizebox{\linewidth}{!}{%
\begin{tabular}{@{}llrrr@{}}
\toprule
Trigger size &
  \begin{tabular}[c]{@{}l@{}}Poisoning\\ rate (\%)\end{tabular} &
  \begin{tabular}[c]{@{}r@{}}Clean data \\ mAP (\%) $\uparrow$\end{tabular} &
  \begin{tabular}[c]{@{}r@{}}Poisoned data \\ mAP (\%) $\downarrow$\end{tabular} &
  ASR (\%) $\uparrow$ \\ \midrule
15x15 & 15 & 88.65 & 3.05  & 95.28 \\
15x15 & 20 & 84.25 & 5.69  & 91.52 \\
20x20 & 15 & 93.17 & 45.34 & 62.44 \\
20x20 & 20 & 89.09 & 47.23 & 84.03 \\ \bottomrule
\end{tabular}%
}
\vspace{-0.5em}
\caption{Performance of BadFusion using different poisoning rates and trigger sizes against MVX-Net fusion method.}
\label{tab:abl_pr_ts}
\vspace{-1.5em}
\end{table}

\vspace{0.2em}
\noindent\textbf{Robustness Analysis.}
The primary focus of our paper is on the attack aspect of fusion-based 3D object detection, and we also conduct a robustness analysis of BadFusion to investigate whether the backdoor will be removed after applying defenses. To the best of our knowledge, no existing defenses are specifically designed to counter the BadFusion attack. We evaluate BadFusion's robustness against three prevalent defense strategies: 1) Input Noising \cite{xu2017feature}, 2) JPEG Compression \cite{dziugaite2016study}, and 3) Regularization \cite{shafieinejad2021robustness}. The results demonstrate BadFusion's robustness, further emphasizing the vulnerability of fusion-based 3D object detection and the importance of our research. Additional details are discussed in the Appendix.

\vspace{-0.5em}
\section{Conclusion}

This paper presents the first analysis of 2D-oriented backdoor attacks against LiDAR-camera fusion for 3D object detection. By analyzing the existing 2D-oriented backdoor attacks, we find that these attacks are ineffective against fusion models due to the sparsity and inconsistency of backdoor triggers introduced during the fusion process. To address these challenges, we propose BadFusion, an innovative fusion-aware backdoor attack against 3D object detection. By maximizing both effective trigger pixels and consistent trigger patterns among different inputs, BadFusion successfully performs backdoor attacks against state-of-the-art LiDAR-camera fusion methods and realizes two attack goals: resizing the bounding boxes and disappearing the objects.
Compared with existing 2D-oriented backdoor attacks, BadFusion achieves a much higher attack success rate and low Poisoned data mAP.  
We hope our analysis will enhance safety awareness for autonomous driving and promote further research in this field.

\section{Acknowledgment}
This work was supported by the National Science Foundation under Grant No. 2426318, 2427316, and 2151238.

\bibliographystyle{named}

\bibliography{ijcai24}

\clearpage
\appendix
\section{Appendix}

\subsection{BadFusion setup.}
In the proposed BadFusion attack, we determine the (x, y) coordinates for trigger insertion based on dense regions in the corresponding 2D LiDAR projections onto image planes. We filter out LiDAR points located within the bounding box of the vehicle intended for poisoning and employ a sliding window to identify the densest region with most LiDAR points. 
For resizing attacks, we reduce the sizes of the 3D Bounding Box by 75\% to facilitate a reduced size bounding box attack. For disappearing attacks, we change the (x, z) coordinate by doubling or halving the values of x and z, i.e., making the vehicles farther or closer distance from the target vehicle.

\subsection{Fusion Methods} 

\noindent\textbf{MVX-Net.} We trained the MVX-Net fusion model \cite{sindagi2019mvx} using Faster R-CNN as backbone. This model integrates a frozen 50-layer ResNet architecture with a Feature Pyramid Network neck in its image backbone to enhance feature representation. Point cloud processing is conducted using a voxelization block, and the Camera and LiDAR features are fused using the PointFusion method. Finally, the fused features are processed with the Dynamic Voxel Feature Encoding module and 3D Region Proposal Network for 3D Object Detection.

\vspace{0.2em}
\noindent\textbf{Focals Conv-F.} We trained the Focals Conv-F fusion model \cite{chen2022focal} utilizing the PVRCNN \cite{shi2020pv} as the base model. The Focals Conv-F fusion model employs the BaseBEVBackbone as the 2D backbone and their proposed Focal Sparse Convolutional Networks for processing the 3D backbone, leveraging image data and BeV mappings. The extracted Camera and LiDAR features undergo point-wise fusion and serve as input for the proposed Focal Sparse Convolutional network. This network determines which input features deserve dilation and adjusts the output shapes dynamically based on the predicted cubic importance. The resulting features are processed by a set of fully connected layers to predict the 3D bounding boxes of the objects in the scene.

\vspace{0.2em}
\noindent\textbf{EPNet.} We trained the EPNet F fusion model \cite{huang2020epnet} utilizing the two-stream Region Proposal Network (RPN) for proposal generation and a refinement network for bounding box refining. In EPNet, the two-stream RPN is composed of a geometric stream and an image stream, which produce the point features and semantic image features, respectively. Finally, the LI-Fusion module establishes the fine-grained point-wise correspondence between LiDAR and camera image data and fuses the point features and semantic image features based on the correspondence generated by the grid generator. The LI-Fusion module adaptively estimates the importance of the image semantic features and fuses them with the point features to enhance the 3D object detection performance.

\subsection{Experimental Results Discussion}
As shown in Table~\ref{tab:maintable}, the existing backdoor attacks (OptimizedTrigger, BadDet, UntarOD) result in a very low ASR and unchanged Poisoned mAP despite performing poisoning, which shows the existing backdoor attacks fail to mislead the fusion models. 
The clean mAP of existing backdoor attacks for the Focals Conv-F fusion method is higher as compared to the clean mAP for the MVX-Net fusion method because the existing backdoor attacks for MVX-Net learned memorization instead of learning poisoned behavior (i.e., irrespective of whether the trigger is present or not they predicted resized (smaller) bounding box), thereby affecting clean mAP. In the Focals Conv-F fusion method, the poisoning does not have any impact on the existing backdoor attacks. So, they behave similarly to the clean model and have higher clean mAP.
 
For the EPNet fusion method, the existing backdoor attack (BadDet) result is consistent with the existing backdoor attack results for Focals Conv-F fusion, where they have a low ASR and unchanged Poisoned mAP. Whereas, the existing backdoor attack (OptimizedTrigger, UntarOD) learned the backdoor behavior learning, as the EPNet fusion method adds continuous image feature segmentation instead of sparse image features during fusion, unlike MVX-Net or Focals Conv-F, resulting in weaker built-in defense against backdoor attacks.

\noindent\textit{Why do the backdoor attacks on EPNet perform better than other fusion methods?} 

We believe the fusion manner proposed in EPNet plays a critical role in its vulnerability to backdoor attacks.
To enhance the correspondence between the camera and LiDAR features, EPNet first uses LiDAR features to guide the image features, named LiDAR-guided Image Fusion (LI-Fusion), and then generates point-wise image features. Here, a bilinear interpolation-based image sampler is applied for each LiDAR point to extract neighboring image features.
The interpolation potentially creates an attack surface - despite the lack of corresponding LiDAR points, the backdoor triggers injected in 2D camera space can still be effective through the neighboring LiDAR points during the interpolation process. This eliminates the sparsity challenges in existing 2D-oriented backdoor attacks (e.g., OptimizedTrigger and UntarOD).
Therefore, as shown in the results, the performance of existing attacks on EPNet is higher than that of the other fusion methods, such as MVX-Net and Focals-ConvF.
Note that our proposed BadFusion attack is effective in various fusion methods (MVX-Net, Focals Conv-F, and EPNet), while the existing attacks are effective only when backdoor triggers are activated in the fusion, e.g., through the interpolation in EPNet.

\textit{Why the triggers in our proposed BadFusion method would remain dense and consistent regardless of fusion methods?}

BadFusion is designed to be fusion-process agnostic by dealing with two general challenges to designing 2D triggers against fusion-based 3D object detection: trigger sparsity and inconsistency. First, to ensure the trigger density, BadFusion identifies the dense area of LiDAR points to place the trigger, which preserves sufficient 2D trigger information in 3D space. Here, the density of LiDAR points is irrelevant to fusion methods. Second, to ensure the consistency of the trigger pattern, BadFusion employs a simple and effective trigger pattern with a uniform solid color, which is also irrelevant to fusion methods. Therefore, the trigger remains dense and consistent for any given fusion method. To verify this, in the paper, we use the same attack (the same poisoned dataset) against two different fusion methods and show that regardless of the fusion methods, the BadFusion attack can mislead the object detection model.

\begin{figure}[!tb]
\centering
\begin{subfigure}[b]{0.45\linewidth}
\centering
\includegraphics[width=0.3\linewidth]{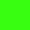}
\caption{uniform solid pattern}
\label{fig:solid}
\end{subfigure}
\begin{subfigure}[b]{0.45\linewidth}
\centering
\includegraphics[width=0.3\linewidth]{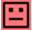}
\caption{almost solid pattern}
\label{fig:almost_solid}
\end{subfigure}
\caption{Different trigger patterns used in BadFusion.}
\label{fig:trigger_pattern}
\vspace{-1em}
\end{figure}

\subsection{Additional Experiments}

\noindent\textbf{Different trigger sizes during inference.}\label{sec:trigger_size} In previous backdoor attacks, the attacker uses the same trigger size in the training and inference phase to keep the same trigger pattern.
In this experiment, we use trigger sizes of $15 \times 15$ in the training and different trigger sizes (from $10\times 10$ to $50\times 50$) in the inference phase. 
Since we use the uniform solid color in the trigger, the increased trigger size does not affect the consistency of the trigger pattern but can potentially increase the number of effective trigger pixels after fusion.
Table~\ref{tab:abl_inferencetriggersize1} shows the results of BadFusion with different attack goals. 
For resizing attacks, increasing trigger sizes in the inference significantly improves the attack performance of BadFusion. For example, with the trigger size of $30\times 30$, the ASR of BadFusion reaches over 97\%. 
This is mainly due to the increased number of effective trigger pixels on 2D LiDAR projection. 
In the meanwhile, we find the increasing trigger size in the inference phase does not affect too much on disappear attacks.
\begin{table*}[!tb]
\centering
\small
\begin{tabular}{@{}lcrrr@{}}
\toprule
Attack Goal & Inference Trigger Size & Clean data mAP (\%) $\uparrow$ & Poisoned data mAP (\%) $\downarrow$ & ASR (\%) $\uparrow$ \\ \midrule
\multirow{5}{*}{Resizing Attack} & 10x10 & \multirow{5}{*}{93.75} & 14.00 & 90.06 \\
        & 20x20    &  & 1.17 & 96.74 \\
        & 30x30    &  & 0.30 & 97.38 \\
        & 40x40   &  & 0.16 & 97.76 \\
        & 50x40   &  & 0.14 & 97.72 \\
\hline
\multirow{5}{*}{Disappear Attack (farther)} & 10x10 & \multirow{5}{*}{93.75} & 22.29 & 92.61 \\
        & 20x20    &  & 2.81 & 89.32 \\
        & 30x30    &  & 0.67 & 88.57 \\
        & 40x40   &  & 0.19 & 88.03 \\
        & 50x40   &  & 0.15 & 87.38 \\
\hline
\multirow{5}{*}{Disappear Attack (closer)} & 10x10 & \multirow{5}{*}{93.75} & 44.53 & 95.96 \\
        & 20x20    &  & 11.37 & 93.89 \\
        & 30x30    &  & 3.39 & 92.91 \\
        & 40x40   &  & 1.57 & 92.40 \\
        & 50x40   &  & 0.77 & 91.42 \\
\bottomrule
\end{tabular}
\caption{Impact of different trigger sizes in the Inference for the proposed BadFusion against MVX-Net under resizing attack, disappear attack (farther), and disappear attack (closer).
}
\label{tab:abl_inferencetriggersize1}
\end{table*}

\begin{figure}[!tb]
\centering
\includegraphics[width=\columnwidth]{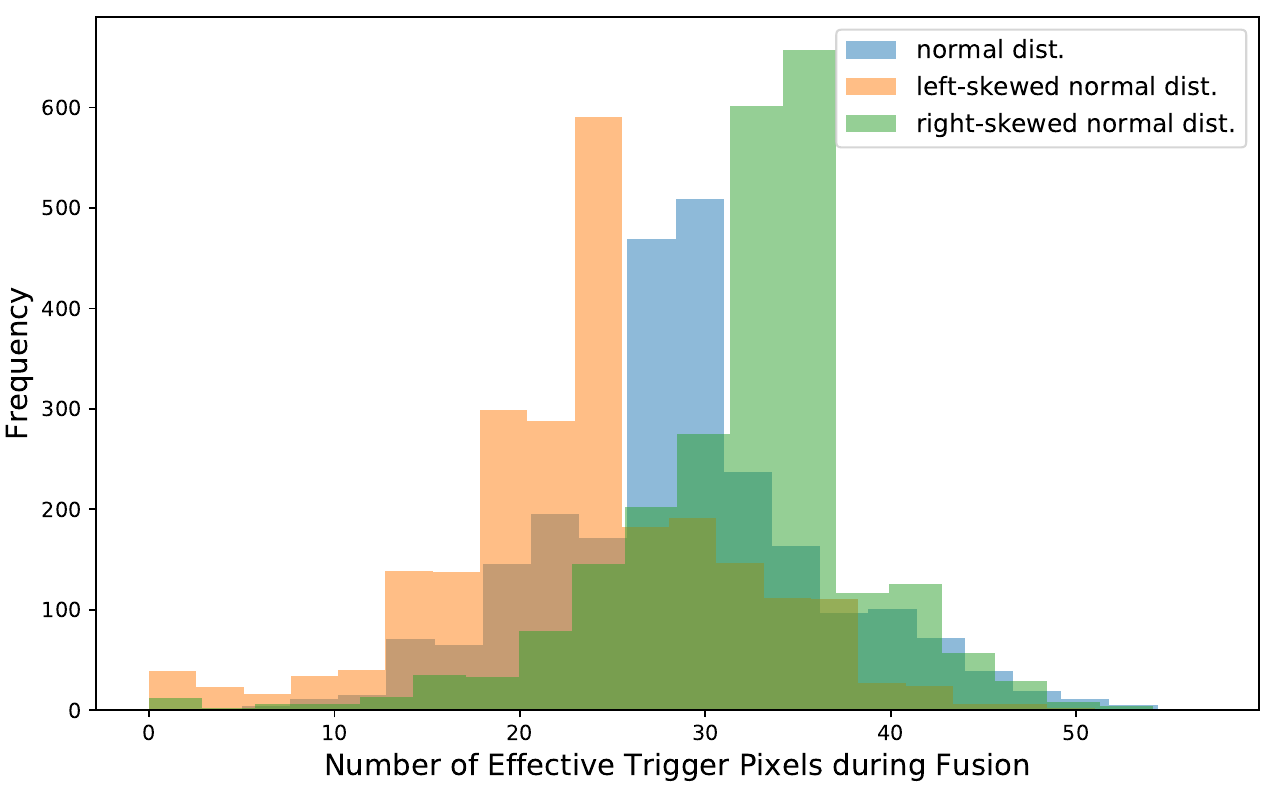} %
\caption{Different distributions of backdoor sample selection in BadFusion against MVX-Net.}
\label{fig:backdoor_selection}
\end{figure}

\vspace{0.2em}
\noindent\textbf{Selection of poisoned samples.}
We investigate the impact of poisoned sample selection on the attack performance. 
We found that the number of effective trigger pixels after fusion plays an important role in backdoor attack performance.
In the main paper, we select poisoned samples whose effective trigger pixels follow Gaussian distributions. In this experiment, we select two additional distributions for poison sample selection: right-skewed distribution and distorted normal distribution, as shown in Figure~\ref{fig:backdoor_selection}.
The other attack setups follow the main paper: we use a trigger size of $15\times 15$ and a poisoning rate of $15$\%. 

We find that the selection of poisoned samples significantly affects the attack performance. As shown in Table~\ref{tab:abl_samples_selection}, the normal distribution shows a much better attack performance than other selections (low Poisoned mAP and high ASR). Any deviation (left-skewed distribution or right-skewed distribution) in the consistency or uniformity of transformed trigger feature pixels across training samples hampers the poisoned model's ability to learn the targeted poisoned behavior. The performance of the poisoned model drops to $87.01$\% for left-skewed distribution and $60.57$\% for right-skewed distribution. 
We believe this is due to the fact that the normal distribution selection better covers different trigger patterns, i.e., different numbers of effective trigger pixels, that may occur in the inference phase. Therefore, the injected trigger can be better ``generalized'' to attacks in the inference.

\begin{table*}[!tb]
\centering
\small
\vspace{1em}
\caption{Impact of switching to emoji trigger on baselines on MVX-Net Fusion Method.
}
\vspace{1em}
\begin{tabular}{@{}lrrr@{}}
\toprule
Backdoor attack & Clean data mAP (\%) $\uparrow$ & Poisoned data mAP (\%) $\downarrow$ & ASR (\%) $\uparrow$ \\ \midrule
OptimizedTrigger & 34.95 & 47.23 & 37.35 \\
BadDet & 29.99 & 41.49 & 45.08 \\
UntarOD & 30.98 & 42.89 & 40.06 \\ \bottomrule
\end{tabular}%
\label{tab:emoji_baselines}
\end{table*}

\begin{table}[!tb]
\centering
\small
\vspace{1em}
\vspace{1em}
\begin{tabular}{@{}lrlr@{}}
\toprule
\begin{tabular}[c]{@{}l@{}}Sample selection\end{tabular} & \begin{tabular}[c]{@{}r@{}}Clean\\ mAP (\%) $\uparrow$ \end{tabular} & \multicolumn{1}{r}{\begin{tabular}[c]{@{}r@{}}Poisoned\\ mAP (\%) $\downarrow$ \end{tabular}} & ASR (\%) $\uparrow$  \\ \midrule
normal dist. & 88.65 & 3.05 & 95.28 \\
left-skewed dist. & 91.88 & 19.44 & 87.01 \\
right-skewed dist. & 40.24 & 42.36 & 60.57 \\ \bottomrule
\end{tabular}%
\caption{Impact of different poison sample selections in BadFusion.}
\label{tab:abl_samples_selection}
\end{table}

\begin{figure*}[!tb]
\centering
\includegraphics[width=0.7\textwidth]{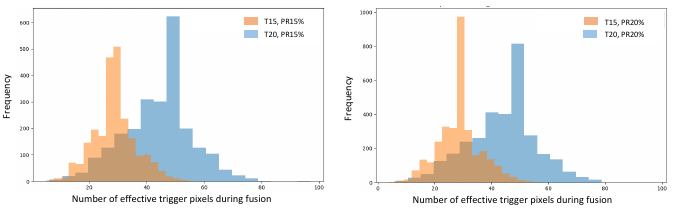} %
\caption{Distribution of effective trigger pixels during fusion using different trigger sizes. We conduct BadFusion attacks against the MVX-Net fusion model to resize attacks. T15 and T20 represents trigger size $15 \times 15$ and $20 \times 20$, respectively. PR15\% and PR20\% represents poisoning rate $15$\% and $20$\%, respectively. }
\label{fig:pr_ts}
\end{figure*}

\vspace{0.2em}
\noindent\textbf{Experiment details on the impact of trigger size and poisoning rate.}

We have found increasing the trigger size does not necessarily improve attack performance. This is mainly due to the effective trigger pixel after transformation being inconsistent with various trigger sizes. Figure~\ref{fig:pr_ts} illustrates that the range of effective trigger pixels for a trigger size of $15 \times 15$ varies from $0$ to $60$, with a primary range of $25$ to $35$. 
In contrast, for a trigger size of $20 \times 20$, the range of effective trigger pixels varies from $0$ to $80$, with a primary range of $38$ to $55$.
The small range of effective trigger pixels for trigger size $15 \times 15$ results in a more consistent trigger pattern across samples in the training and inference phase, thereby leading to a better attack performance for a smaller trigger size ($15 \times 15$). We will clarify this in our paper.

Additionally, the BadFusion poisoned model with a poisoning rate $15$\% and $15 \times 15$ trigger size outperforms the poisoning rate $20$\% poisoned model. We think this is because the increment of $5$\% data (shifting from $15$\% poisoning rate to $20$\% poisoning rate) only contributes an additional 342 samples within the consistent number of effective trigger pixels range while introducing 458 samples within the inconsistent number of effective trigger pixels range. This imbalance disrupts the model's overall learning process. Conversely, the $20 \times 20$ trigger size and poisoning rate $20$\% poisoned model surpasses the $15$\% poisoning rate model. Here, the additional $5$\% data (shifting from $15$\% poisoning rate to $20$\% poisoning rate) contributes 443 samples within the consistent number of effective trigger pixels range and only 357 within the inconsistent number of effective trigger pixels range, thereby facilitating the model's learning process.

\vspace{0.2em}
\noindent\textbf{Impact of switching to an almost solid trigger pattern on baselines.}

Table~\ref{tab:emoji_baselines} presents the impact of switching to an almost solid trigger pattern on the MVX-Net Fusion Method compared to the baselines (existing backdoor attacks). The results align with the main findings presented in Table~\ref{tab:maintable}, where the existing backdoor attacks (OptimizedTrigger, BadDet, UntarOD) prove ineffective in manipulating the fusion detector's predictions, even when employing an almost solid trigger pattern for the backdoor attack. Conversely, the effectiveness of BadFusion with a more stealthy, almost solid trigger pattern is demonstrated in Table~\ref{tab:abl_realtrigger}.

\vspace{0.2em}
\noindent\textbf{Robustness Analysis.}

We evaluate BadFusion's robustness against three prevalent defense strategies: 1) Input Noising \cite{xu2017feature}, 2) JPEG Compression \cite{dziugaite2016study}, and 3) Regularization \cite{shafieinejad2021robustness}.

Input Noising, as described in \cite{xu2017feature}, involves introducing Gaussian noise with a mean of zero and a specified standard deviation to the entire image. On the other hand, JPEG Compression, as discussed in \cite{dziugaite2016study}, employs a process similar to input reconstruction. However, instead of utilizing an autoencoder, the image is compressed using the JPEG compression algorithm. Both of these methods work by modifying the poisoned images in a way that damages the consistency of the backdoor trigger, degrading the effectiveness of the backdoor attack. In our experiments for Input Noising, we add varying noise levels (0 to 10 out of 255) to each pixel of the camera input data. For JPEG Compression, we compress the images to retain 60\% of their original quality. We performed training and evaluation of the corresponding poisoned model with a Resizing attack goal on the MVX-Net fusion model using the experimentation settings stated in the Evaluation settings of our main paper. The results indicate that even with Input Noising and JPEG Compression, the attack success rate of the poisoned model remained consistently high at 94.94\% and 94.98\%, respectively, with a clean mAP of 89.29\% and 89.95\%.

Regularization, as outlined in \cite{shafieinejad2021robustness}, follows a two-phase approach. In the initial phase, the model undergoes strong regularization, resulting in a reduction in test accuracy. This reduction is later addressed by fine-tuning the model during the second phase. The primary objective of the regularization phase is to shift the model's parameters significantly away from their original values, while the subsequent fine-tuning phase seeks to identify a distinct local minimum. In our experiments for Regularization, we first use a large regularization term (l2 weight decay = 0.5) and then fine-tune the model with a standard regularization. The results show that the attack success rate of the poisoned model remained high at 93.05\% and clean mAP at 87.22\% with Regularization.
Thus, the results demonstrate the robustness of BadFusion, highlighting its effectiveness against fusion-based 3D object detection.

\noindent\textbf{Possible failure cases of proposed BadFusion attack}
Our analysis has identified failure cases arising from two main issues: 1) \textit{Sparse trigger locations:} instances where the LiDAR projection lacks sufficient LiDAR points to transform an adequate number of trigger points, even after identifying a dense location. 2) \textit{Overlapping trigger locations:} In our proposed BadFusion method, we insert the trigger at the densest LiDAR point locations during the poisoning process. However, during inference, if the trigger is placed in a region where the bounding boxes of two vehicles overlap, the poisoned model might either predict a reduced-size bounding box or no detection for this unintended neighboring vehicle, assuming the trigger is present for that vehicle. A potential solution to mitigate this challenge is to avoid inserting triggers in areas of bounding box overlap during the inference phase.

\subsection{Comparison between LiDAR-aware and LiDAR-Free BadFusion attacks}
\begin{figure*}[!tb]
\centering
\includegraphics[width=0.7\textwidth]{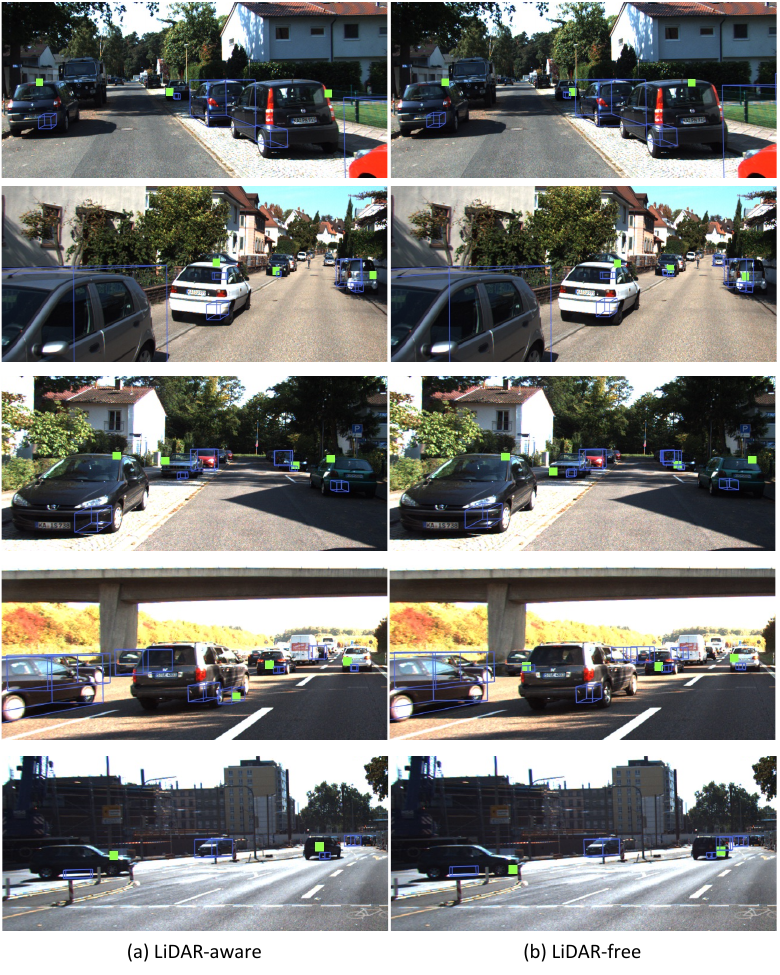} %
\caption{Comparison between LiDAR-aware and LiDAR-free BadFusion.}
\label{fig:lidaraware_lidarfree}
\end{figure*}

Table~\ref{tab:maintable} shows that LiDAR-free BadFusion achieves comparable performance with LiDAR-aware BadFusion, where we assume the attack has access to the LiDAR signals in the inference phase.
Here, we present the example triggers injected by LiDAR-aware and LiDAR-Free BadFusion attacks in Figure~\ref{fig:lidaraware_lidarfree}.
Even though the LiDAR-Free approach does not position the trigger at the vehicle's densest regions, the chosen trigger locations are still effective for backdoor attacks. 
This suggests the effectiveness of the dense region detector in BadFusion.

\subsection{Examples of BadFusion with different attack goals}
\begin{figure*}[!tb]
\centering
\includegraphics[width=0.95\textwidth]{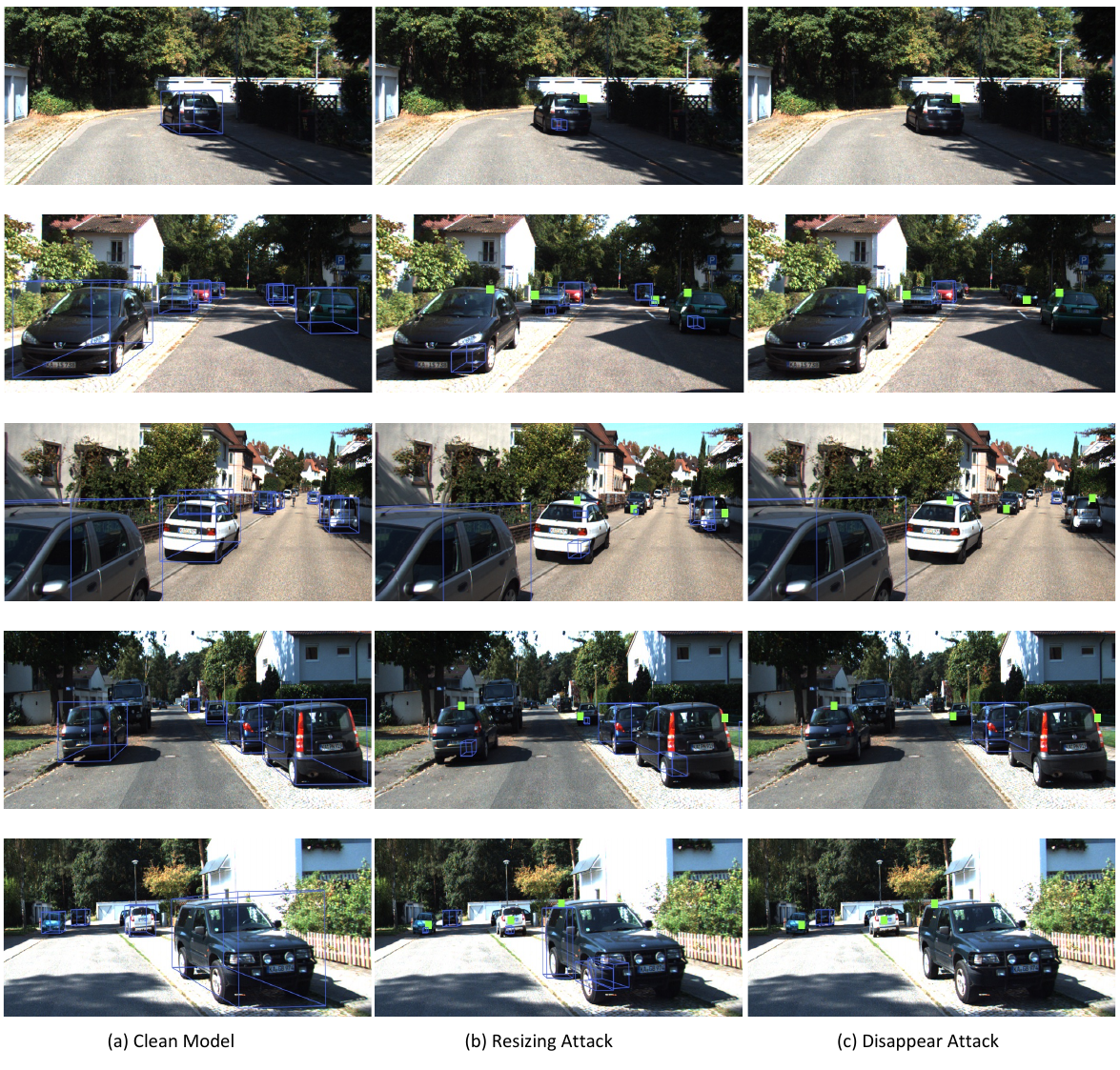} %
\caption{Detection results of the clean model and BadFusion with different attack goals.}
\label{fig:detection_results}
\end{figure*}

In this section, we present examples of BadFusion attacks achieving different attack goals.
Figure~\ref{fig:detection_results} illustrates the detection results of the Clean Model and poisoned model under resizing attacks and disappearing attacks. These comparisons demonstrate the effectiveness of the proposed BadFusion across different attack goals.

\end{document}